\pdfoutput=1

\documentclass[11pt]{article}

\usepackage[preprint]{acl}

\usepackage{times}
\usepackage{latexsym}

\usepackage[T1]{fontenc}

\usepackage[utf8]{inputenc}

\usepackage{microtype}

\usepackage{inconsolata}

\usepackage{graphicx}

\usepackage{hyperref}
\usepackage{stfloats}
\usepackage{booktabs}
\usepackage{multirow}
\usepackage{multicol}
\usepackage{makecell}
\usepackage{array}
\usepackage{amsmath}
\usepackage{diagbox}
\usepackage{tcolorbox}
\usepackage[ruled,linesnumbered]{algorithm2e}
\usepackage{colortbl}
\usepackage{enumitem}
\usepackage{bbm}

\title{From Trade-off to Synergy: A Versatile Symbiotic Watermarking Framework for Large Language Models}

\author{Yidan Wang\textsuperscript{1,2}, Yubing Ren\textsuperscript{1,2}\thanks{Corresponding Author.}, Yanan Cao\textsuperscript{1,2}, Binxing Fang\textsuperscript{3}\\
\textsuperscript{1}Institute of Information Engineering, Chinese Academy of Sciences, Beijing, China \\
\textsuperscript{2}School of Cyber Security, University of Chinese Academy of Sciences, Beijing, China \\
\textsuperscript{3}Hainan Province Fang Binxing Academician Workstation, Hainan, China\\
\texttt{\{wangyidan, renyubing\}@iie.ac.cn}\\
} 

\begin{document}
\maketitle
\begin{abstract}
The rise of Large Language Models (LLMs) has heightened concerns about the misuse of AI-generated text, making watermarking a promising solution. Mainstream watermarking schemes for LLMs fall into two categories: logits-based and sampling-based. However, current schemes entail trade-offs among robustness, text quality, and security. To mitigate this, we integrate logits-based and sampling-based schemes, harnessing their respective strengths to achieve synergy. In this paper, we propose a versatile symbiotic watermarking framework with three strategies: serial, parallel, and hybrid. The hybrid framework adaptively embeds watermarks using token entropy and semantic entropy, optimizing the balance between detectability, robustness, text quality, and security. Furthermore, we validate our approach through comprehensive experiments on various datasets and models. Experimental results indicate that our method outperforms existing baselines and achieves state-of-the-art (SOTA) performance. We believe this framework provides novel insights into diverse watermarking paradigms. Our code is available at \href{https://github.com/redwyd/SymMark}{https://github.com/redwyd/SymMark}.

\end{abstract}

\section{Introduction}
The exceptional capabilities of large language models (LLMs) \cite{touvron2023llama, zhang2022opt} have revolutionized various fields, including creative content generation and automated writing, etc. The widespread accessibility of LLMs has significantly reduced the barriers to using AI-generated content, enabling broader adoption across diverse domains. While this democratization of technology brings substantial benefits, it also introduces critical challenges, including the potential misuse of LLMs for generating malicious content, violating intellectual property rights, and spreading disinformation \cite{liu2024survey}. To address these risks, watermarking has emerged as a promising solution for ensuring the traceability, authenticity, and accountability of LLM-generated content. By embedding invisible identifiers within generated text, watermarking provides a robust mechanism to trace content origins and mitigate misuse.
\begin{figure}[tp]
\centering
\includegraphics[width=\linewidth]{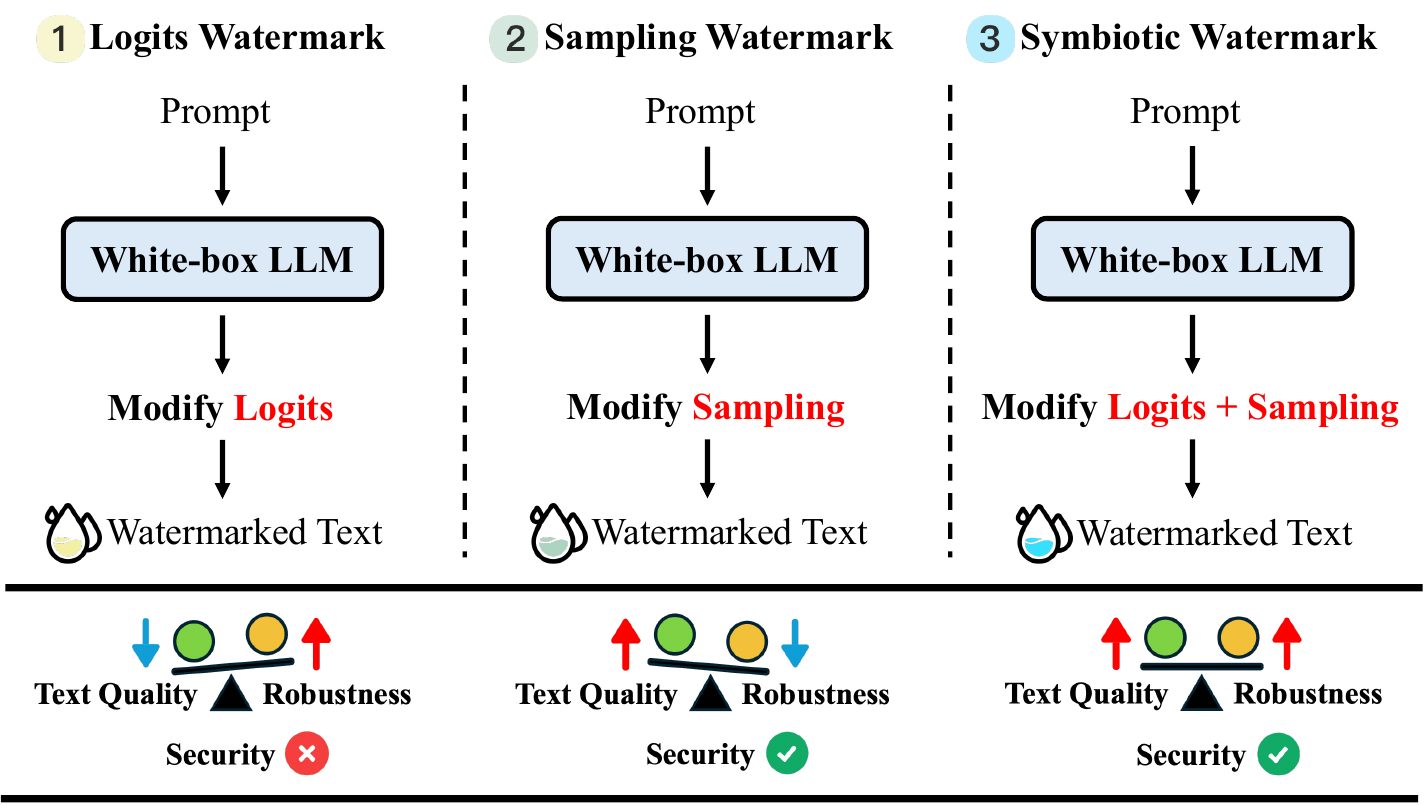}
\caption{Paradigm comparison between our symbiotic watermark framework SymMark and existing logits-based watermark / sampling-based watermark.}
\label{figure 1}
\end{figure}

However, existing watermarking methods face fundamental limitations that hinder their effectiveness in diverse and adversarial scenarios \cite{pmlr-v202-kirchenbauer23a, kuditipudi2024robust}. A key challenge lies in balancing \textbf{robustness} and \textbf{text quality}—increasing watermark strength often compromises the fluency and diversity of generated text while prioritizing quality can weaken robust to adversarial attacks \cite{wu2023dipmark, zhao2024provable, dathathri2024scalable}.  Moreover, the security of watermarks remains a pressing issue. Current methods, such as the KGW family, are vulnerable to attacks like watermark stealing, where adversaries can potentially reverse-engineer watermark rules via frequency analysis, undermining their effectiveness  \cite{jovanovic2024watermark, pang2024no, wu-chandrasekaran-2024-bypassing}. Finally, as shown in Figure \ref{figure 1}, the field lacks golden design principles, as both logits-based and sampling-based watermarkings face inherent trade-offs.

Can robustness, text quality, and security be harmonized to work together, rather than being treated as conflicting objectives? Drawing inspiration from \textit{symbiosis} in natural ecosystems, where different entities coexist and thrive through mutual benefits, we explore a novel perspective for watermarking. We introduce \textbf{SymMark}, a versatile symbiotic watermarking framework that transcends the traditional trade-offs in watermarking design. By transforming these trade-offs into synergy, SymMark combines the strengths of logits-based and sampling-based watermarking, providing an innovative solution that ensures robustness, text quality, and security, even under adversarial conditions.

Building on this symbiotic perspective, SymMark explores three strategies to integrate logits-based and sampling-based watermarking. Serial Symbiotic Watermarking \textbf{(Series)} embeds both watermarks in each token, ensuring high detectability. However, overly strong watermarks can degrade text quality. Parallel Symbiotic Watermarking \textbf{(Parallel)} alternates between the two methods at the token level, balancing robustness and text quality. Yet, it lacks flexibility, unable to adaptively select the optimal watermarking strategy for each token.
To address these issues, we introduce Hybrid Symbiotic Watermarking \textbf{(Hybrid)}, our primary configuration. Hybrid applies a non-linear combination of both watermarking methods, adaptively choosing the most suitable strategy for each token. This may involve applying both watermarks, only one, or skipping watermarking altogether, depending on the token’s context. By dynamically selecting the best strategy based on token and semantic entropy \cite{Shannon, farquhar2024detecting}, Hybrid enhances watermark security, resilience, and fluency. Additionally, we propose a unified algorithm to detect all three strategies effectively and efficiently.


Extensive experiments across multiple datasets and models consistently reveal that SymMark outperforms existing baselines. Specifically, the Serial excels in detectability and robustness, while the Parallel preserves high text quality without weakening watermark strength. Hybrid integrates the strengths of both approaches, making it the most comprehensive and effective strategy.
Our main contributions are as follows:
\begin{itemize}[leftmargin=*]

    \item We systematically explore the integration of logits-based and sampling-based watermarking methods, pioneering a comprehensive approach to their synergy.
    
    \item We propose a versatile symbiotic watermarking framework, \textbf{SymMark}, which incorporates three distinct strategies: Series, Parallel, and Hybrid.

    \item Our exhaustive experiments demonstrate that the SymMark framework achieves state-of-the-art (SOTA) performance in terms of detectability, robustness, text quality, and security.

\end{itemize}

\section{Related Work}
The current mainstream LLM watermarking during the generation stage can be categorized into logits-based and sampling-based.	

\paragraph{Logits-based Watermarking.} 
The pioneering KGW method \cite{pmlr-v202-kirchenbauer23a} uses a hash key to divide the vocabulary into red and green lists, favoring green tokens in the output. To enhance watermark robustness, Unigram \cite{zhao2024provable} introduces a fixed red-green vocabulary partitioning scheme. \citet{ren-etal-2024-subtle} incorporate the vocabulary’s prior distribution, and \citet{ren-etal-2024-robust, he-etal-2024-watermarks, liu2024a, liu2024adaptivetextwatermarklarge, huo2024tokenspecific, fu2024watermarking, chen-etal-2024-watme} determine logits partitioning using semantic embeddings. \citet{hu2024unbiased, wu2024a} explore unbiased watermarking to ensure identical expected distributions between watermarked and non-watermarked texts. To improve watermarked text quality, SWEET \cite{lee-etal-2024-wrote}, EWD \cite{lu-etal-2024-entropy}, and \citet{wouters2023optimizing} optimize watermarking from an entropy perspective. Furthermore, \citet{guan-etal-2024-codeip, fernandez2023bricksconsolidatewatermarkslarge, wang2024towards, yoo-etal-2024-advancing} investigate multi-bit watermarks to obtain higher capacity and convey more information.




\paragraph{Sampling-based Watermarking.} 
In token-level sampling watermarking, \citet{christ2024undetectable} employ a pseudo-random number to guide token generation, though it is unsuitable for real-world LLMs. Meanwhile, AAR \cite{Aaronson} utilizes exponential minimum sampling to embed the watermark, while \citet{fu-etal-2024-gumbelsoft, kuditipudi2024robust} build on this method to enhance text diversity further.	\citet{zhu-etal-2024-duwak} advocate contrastive decoding for sampling, and \citet{dathathri2024scalable} devise a tournament sampling scheme that preserves text quality while ensuring high detection accuracy. In sentence-level sampling watermarking, SemStamp \cite{hou-etal-2024-semstamp} divides the semantic space into watermarked and non-watermarked regions using locality-sensitive hashing. k-SemStamp \cite{hou-etal-2024-k} further optimizes this process with a K-means clustering \cite{macqueen1967some} algorithm. 


\section{Preliminary}
\begin{figure*}[htbp]
\centering
\includegraphics[width=\textwidth]{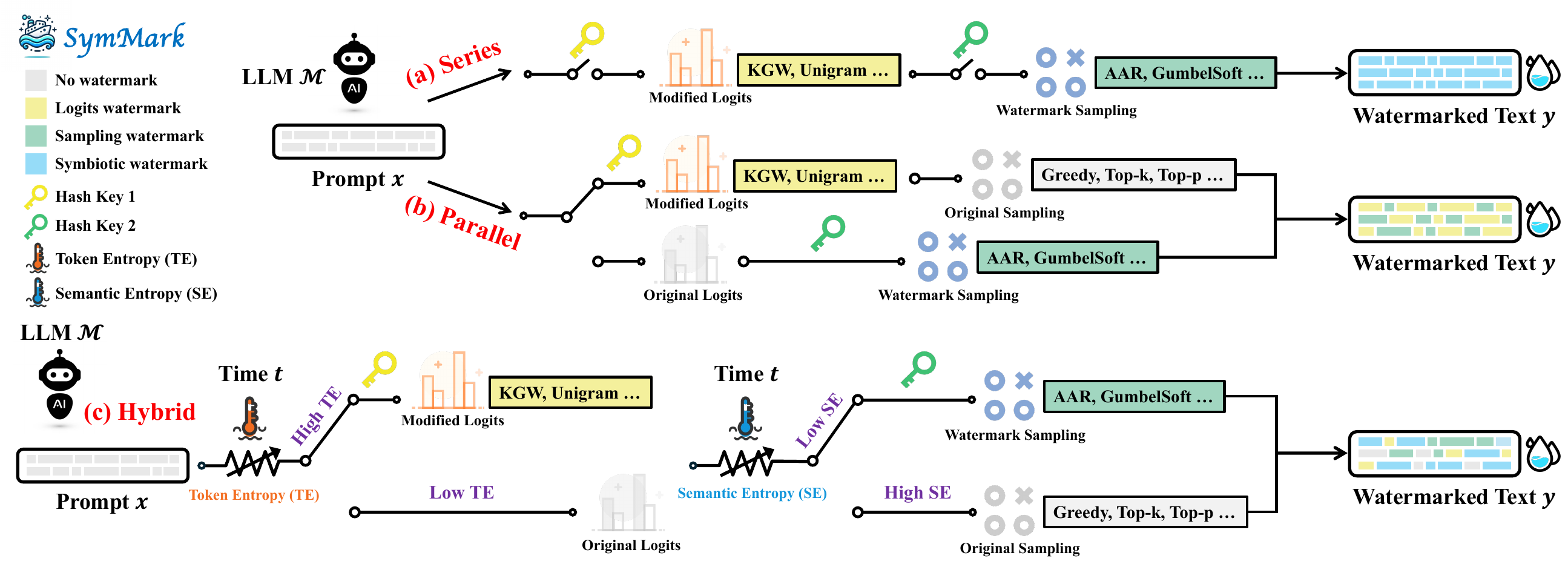}
\caption{A Versatile Symbiotic Watermark Framework for LLMs.}
\label{figure 2}
\end{figure*}

\subsection{LLM Generation}
LLM $\mathcal{M}$ is a transformer-based \cite{vaswani2017attention} autoregressive neural network, characterized by its vocabulary $\mathcal{V}$ and parameters $\theta$. The generation process of $\mathcal{M}$ involves two steps: \textbf{(1)} given prompt $x$ and the previously generated tokens $y_{<t}=\{y_1,...,y_{t-1}\}$, calculate $t$-th token's logits vector $l_t=\mathcal{M}(\cdot\mid x, y_{<t})$ of length $|\mathcal{V}|$, and then normalize it through softmax function to obtain a probability vector $p_t=\operatorname{softmax}(l_t)$; \textbf{(2)} Sample the $t$-th token based on $p_t$. Common sampling methods include greedy search, beam search, and multinomial sampling, among others.

\subsection{LLM Watermarking}
LLM watermarking is embedded into the token generation process by modifying one of two stages: (i) the logits generation stage, or (ii) the sampling stage. A typical watermarking in the logits stage is \textbf{KGW} \cite{pmlr-v202-kirchenbauer23a}, which partitions the vocabulary into red and green lists with the $\gamma$ ratio. This is achieved by hashing the previous $k$ tokens with the watermark key $\xi$ and applying a $\delta$ bias to the logits of each token in the green list, making the LLM more inclined to generate these tokens. During detection, hypothesis testing can determine if the text of length $L$ contains a watermark by analyzing the number of green list tokens $n_\text{green}$. Specifically, if the proportion of green tokens significantly exceeds $\gamma$, with a high $\text{z-score}=(n_\text{green}-\gamma L)/\sqrt{L\gamma(1-\gamma})$ above the threshold, the text is considered watermarked. 
\citet{zhao2024provable} propose \textbf{Unigram}, a robust variant of KGW, that utilizes a fixed global split between red and green lists to generate watermark logits. However, Unigram is susceptible to statistical analysis, which could reveal the tokens classified as green.
In contrast, the watermark in the sampling stage avoids altering the logits and embeds the watermark by modifying the sampling algorithm. \textbf{AAR} \cite{Aaronson} proposes an exponential scheme to select tokens using $y_t = \arg\max_{i\in\mathcal{V}}(r_t^i)^{1/p_t^i}$, where $r_t\in[0,1]^{|\mathcal{V}|}$ is a random sequence, obtained by hashing the previous $h$ tokens with a fixed watermark key $\xi$ or by shifting the watermark key \cite{kuditipudi2024robust} to get multiple random sequences $r=\xi^{(1)},...,\xi^{(m)}$. During detection, if the hash scores $r_t$ of the tokens in the observed sequence are high, the $p$-value will be low, indicating the presence of a watermark.

\section{SymMark}

This section first introduces three symbiotic watermark strategies—Series, Parallel, and Hybrid. Then outlines a unified symbiotic watermark detection algorithm.

\subsection{Series Symbiotic Watermark}\label{4.1} 
To fully embed the two watermarks and maximize the watermark signal, we designed the series symbiotic watermark, as illustrated in Figure \ref{figure 2} (a). When LLM generates $t$-th token, we first apply a logits-based watermarking $\mathcal{A}_w$ (e.g., KGW, Unigram, etc.) to modify the logits distribution $l_t$, followed by normalization via softmax function. During the sampling stage, we employ a sampling-based watermarking $\mathcal{S}_{w}$ (e.g., AAR, EXP, etc.) to generate the current token $y_t$:

\begin{equation}
    y_t=\mathcal{S}_{w}(\operatorname{softmax}(\mathcal{A}_w(l_t)))
\end{equation}

\subsection{Parallel Symbiotic Watermark}\label{4.2}
To independently embed two watermark signals while minimizing their mutual interference, we propose a parallel symbiotic watermark, as shown in Figure \ref{figure 2} (b). This approach embeds either a logits-based or sampling-based watermark as the LLM generates the current token $y_t$. Specifically, at odd positions, the logits-based watermarking $\mathcal{A}_w$ modifies the logits distribution to embed the watermark, preserving the original sampling algorithm $\mathcal{S}_o$. At even positions, the logits distribution remains unchanged, embedding the watermark with the sampling-based watermarking $\mathcal{S}_w$. The formal representation is as follows, where $k \in \mathbbm{N}$:

\begin{equation}
    y_t = 
    \begin{cases}
    \mathcal{S}_o(\operatorname{softmax}(\mathcal{A}_w(l_t))), & t=2k \\
    \mathcal{S}_{w}(\operatorname{softmax}(l_t)), & t=2k+1 \\
    \end{cases}
\end{equation}

\SetKwInput{KwParams}{Params}

\begin{algorithm}[tp]
\footnotesize
  \SetAlgoLined
  \KwIn{$\text{LLM }\mathcal{M},\text{prompt } x, \text{ComputeEntropy }\mathcal{E}$}
  \KwParams{\mbox{Length $T, \text{TE Threshold } \alpha, \text{SE Threshold } \beta$}}
  \KwOut{Watermarked Text $y_{1:T}$}
  \For{$t=1,2...,T$}{
    $l_t\leftarrow\mathcal{M}(x,y_{<t})$
    
    $\hat{l}_t\leftarrow l_t$

    \tcp{\textcolor{gray}{\mbox{Compute Two Entropy}}}
    $H_{TE}, H_{SE} \leftarrow \mathcal{E}(l_t)$     
    
    \tcp{\textcolor{gray}{Add Logits Watermark}}
    \If{$H_{TE} > \alpha$}{
        $\hat{l}_t\leftarrow\mathcal{A}_w(l_t)$
    
    }
    
    $\hat{p}_t\leftarrow\operatorname{softmax}(\hat{l}_t)$

    \tcp{\textcolor{gray}{Add Sampling Watermark}}
    \ElseIf{$H_{SE} < \beta$}{
    
        $y_t\sim S_w(\hat{p}_t)$
        
        \textbf{continue}
        
    }
    \tcp{\textcolor{gray}{Origin Sampling Method}}
    $y_t\sim S_o(\hat{p}_t)$
  }
 \caption{\mbox{Hybrid Symbiotic Watermark}}
 \label{Algorithm 1}
\end{algorithm}

\subsection{Hybrid Symbiotic Watermark}\label{4.3}

To achieve a synergy between logits-based and sampling-based watermarks, we propose an adaptive hybrid symbiotic watermarking method, as illustrated in Figure \ref{figure 2} (c). This approach leverages two key entropy measures to dynamically decide the watermarking strategy: token entropy determines whether to apply logits-based watermarking, while semantic entropy governs the use of sampling-based watermarking.

\definecolor{denim}{rgb}{0.08, 0.38, 0.74}


\paragraph{Token Entropy} 
Derived from Shannon entropy \cite{Shannon}, quantifies the uncertainty in the logits distribution of a token at the current time step $t$. Given the model’s logits output, we apply softmax normalization to obtain the probability $p_t^i$ for each token $i\in\mathcal{V}$, and compute token entropy as follows:
\begin{equation}
    H_{TE}=-\sum_ip_t^i\log p_t^i, \quad i\in\mathcal{V} \label{equation 3}
\end{equation}

Token entropy serves as the basis for applying logits watermarking because it reflects the model’s confidence in generating a particular token. \textbf{Low token entropy} (high confidence) indicates the model strongly prefers a specific token, meaning that altering logits may significantly affect the fluency and naturalness of the generated text. Thus, applying logits watermarking could be intrusive. \textbf{High token entropy} (low confidence) indicates the model exhibits greater uncertainty, with multiple competing candidates in the logits distribution. Since the token choice is inherently unstable, modifying logits introduces minimal disruption to text quality while ensuring effective watermark embedding.  

\begin{figure}[tp]
\centering
\includegraphics[width=\linewidth]{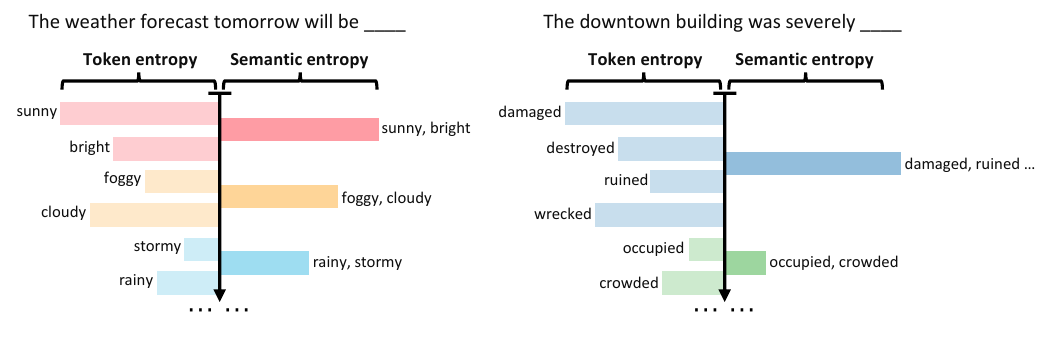}
\caption{High Token Entropy with High Semantic Entropy (Left) and Low Semantic Entropy (Right).}
\label{figure 3}
\end{figure}

\paragraph{Semantic Entropy} 
Semantic entropy measures the diversity of the top-$k$ candidate tokens at time step $t$ in terms of their semantic meaning. To compute semantic entropy, we extract the embeddings of the top-$k$ tokens from the logits distribution and cluster them into $n$ groups $\mathcal{C}=\{\mathcal{C}_1,..., \mathcal{C}_n\}$ using K-means \cite{macqueen1967some}. The logits are then merged according to the cluster assignments, as shown in Equation \ref{equation 4}, and the final semantic entropy is computed from the merged logits, as detailed in Equation \ref{equation 5}. 
   
\begin{align}
q_t^j=\sum_{i=1}^{|\mathcal{C}_j|}p_t^{i}&, \quad i\in\mathcal{C}_j\label{equation 4}\\
H_{SE}=-\sum_jq_t^j\log q_t^j&, \quad j\in\{1,...,n\}\label{equation 5} 
\end{align}


Semantic entropy determines whether to apply sampling watermarking by assessing how semantically diverse the top-ranked candidates are. As illustrated in Figure \ref{figure 3}, \textbf{low semantic entropy} (high semantic similarity) means that the top candidates have similar meanings, implying that replacing one with another will have a negligible impact on text interpretation. Thus, adding a sampling watermark is unlikely to alter the meaning of the generated content. While \textbf{high semantic entropy} (low semantic similarity) indicates the top candidates exhibit substantial semantic variation. In such cases, altering the sampling process could disrupt the intended meaning of the sentence, making sampling watermarking undesirable. Experimental analysis is provided in Appendix \ref{The impact of Semantic Entropy}.



Algorithm \ref{Algorithm 1} details the overall process. Given a logits distribution generated by the LLM $\mathcal{M}$, we first compute token entropy $H_{TE}$ and semantic entropy $H_{SE}$. If $H_{TE}$ exceeds the predefined threshold $\alpha$, logits watermarking is applied; otherwise, the logits remain unchanged. After normalization via softmax and sampling, we check $H_{SE}$: if it falls below the predefined threshold $\beta$, sampling watermarking is applied, ensuring that the final text preserves semantic integrity. This hybrid strategy dynamically selects the optimal watermarking method for each token, achieving robust and high-quality watermark embedding.

\begin{algorithm}[tp]
\footnotesize
  \SetAlgoLined
  \KwIn{$\mathcal{M}, y_{1:T},\mathcal{D}_{l}, \mathcal{D}_{s}, z_1, z_2$}
  \KwOut{$I\text{: True (Watermarked) or False}$}

  $I_l\leftarrow \text{False}$
  
  $I_s\leftarrow \text{False}$

  \tcp{\textcolor{gray}{\mbox{Logits Watermark Detection}}}\If{$\mathcal{D}_l(\mathcal{M},y_{1:T})>z_1$}{$I_l\leftarrow \text{True}$} 
  
  \tcp{\textcolor{gray}{\mbox{Sampling Watermark Detection}}}\If{$\mathcal{D}_s(\mathcal{M},y_{1:T})>z_2$}  {$I_s\leftarrow \text{True}$}

  $I\leftarrow I_l \mid I_s$
 \caption{\mbox{Symbiotic Watermark Detection}}
 \label{Algorithm 2}
\end{algorithm}

\subsection{Symbiotic Watermark Detection}\label{4.4}

Algorithm \ref{Algorithm 2} presents the symbiotic watermark detection process. Given the watermark model $\mathcal{M}$, the generated content $y_{1:T}$, the logits-based detection algorithm $D_l$, and the sampling-based detection algorithm $D_s$, the watermark is deemed present if any watermark signal is detected due to the method’s low false positive rate. Theoretically, tokens can be grouped according to different symbiotic watermark frameworks for detection. Further analysis is provided in Appendix \ref{Group-based Detection}.

\section{Experimental Setup}
\definecolor{logits}{RGB}{243, 240, 166}
\definecolor{sampling}{RGB}{172, 212, 193}
\definecolor{symbiotic}{RGB}{165, 218, 245}

\paragraph{Dataset and Prompt.} To measure detectability, we follow \citet{pmlr-v202-kirchenbauer23a, zhao2024provable} and use subsets of the news-like C4 dataset \cite{raffel2020exploring} and the long-form OpenGen dataset \cite{krishna2023paraphrasing} to insert watermarks. For each sample, the last 200 tokens are treated as natural text (i.e., human-written), while the remaining tokens from the start are used as prompts. We then generate $T=200 \pm30$ tokens (i.e., watermarked text) using LLMs conditioned on the prompts. To evaluate text quality, we followed the Waterbench \cite{tu-etal-2024-waterbench} framework and tested four downstream tasks: Factual Knowledge, Long-form QA, Code Completion, and Text Summarization. Details are in Appendix \ref{Downstream Task Datasets}.

\paragraph{Models.} We conducted experiments using three model series: the OPT series (OPT-6.7B, OPT-2.7B, OPT-1.3B) \cite{zhang2022opt}, the LLaMA series (LLaMA3-8B-Instruct, LLaMA2-7B-chat-hf) \cite{dubey2024llama, touvron2023llama}, and the GPT series (GPT-J-6B) \cite{gpt-j}. Notably, semantic clustering requires using a model with the same tokenizer as the original watermark model.

\paragraph{Baselines.} We compared SymMark with dozens of existing methods, including logits-based watermark KGW \cite{pmlr-v202-kirchenbauer23a}, Unigram \cite{zhao2024provable}, SWEET \cite{lee-etal-2024-wrote}, EWD
\cite{lu-etal-2024-entropy}, DIP
\cite{wu2024a}, Unbiased \cite{hu2024unbiased} and sampling-based watermark AAR \cite{Aaronson}, EXP \cite{kuditipudi2024robust}, ITS \cite{kuditipudi2024robust}, GumbelSoft
\cite{fu-etal-2024-gumbelsoft}, SynthID
\cite{dathathri2024scalable}. Detailed introductions are in Appendix \ref{Baseline Settings}.

\paragraph{Evaluation Metrics.} 
Watermark detectability is evaluated using True Positive Rate (TPR), True Negative Rate (TNR), Best F1 Score, and AUC metrics. Watermark robustness is assessed through the AUROC curve, which illustrates the FPR (False Positive Rate) and TPR across varying thresholds. 

\paragraph{Implementation Details.}
Our symbiotic watermark selects the representative logits-based Unigram watermark \cite{zhao2024provable}, with the classic sampling-based AAR watermark \cite{Aaronson}. The hybrid symbiotic watermark employs the K-means \cite{macqueen1967some} clustering algorithm with the following default hyperparameters: Top-$k$ token numbers $k=64$, clusters number $n=10$, token entropy threshold $\alpha=1.0$, and semantic entropy threshold $\beta=0.5$. Detailed Hyperparameter Analysis is in Appendix \ref{Hyperparameter Analysis}.

\section{Experimental Analysis}
To demonstrate SymMark's superiority, we evaluated it in four aspects: detectability, robustness, text quality, and security. The experimental results show that the Serial excels in detectability and robustness, Parallel better preserves text quality, and Hybrid achieves the best overall balance.

\begin{table*}[tp]
    \centering
    \resizebox{\textwidth}{!}{\begin{tabular}{ccccccccccccccccc}
    \toprule[1.5pt]
    \multirow{3}{*}{Watermark} & \multicolumn{8}{c}{\textsc{C4 Dataset}} & \multicolumn{8}{c}{\textsc{OpenGen Dataset}}\\

    & \multicolumn{4}{c}{\textsc{OPT-6.7B}} & \multicolumn{4}{c}{\textsc{GPT-J-6B}} & \multicolumn{4}{c}{\textsc{OPT-6.7B}} & \multicolumn{4}{c}{\textsc{GPT-J-6B}}\\
    
    \cmidrule(lr){2-5}
    \cmidrule(lr){6-9} \cmidrule(lr){10-13}
    \cmidrule(lr){14-17}
    
    & TPR & TNR & F1 & AUC & TPR & TNR & F1 & AUC & TPR & TNR & F1 & AUC & TPR & TNR & F1 & AUC \\
    \midrule


    
    \rowcolor{logits!30}\multicolumn{17}{c}{{ \textbf{Logits Watermark}}} \\
    KGW & 
    0.990 & 1.000 & 0.994 & 0.999 & 0.995 & 0.995 & 0.995 & 0.999 & 1.000 & 1.000 & 1.000 & 1.000 & 0.995 & 0.990 & 0.992 & 0.997 \\
    DIP & 
    0.985 & 0.995 & 0.989 & 0.999 & 0.990 & 1.000 & 0.994 & 0.995 & 0.995 & 0.995 & 0.995 & 0.998 & 0.940 & 0.995 & 0.966 & 0.985 \\
    EWD & 
    0.995 & 0.995 & 0.995 & 0.997 & 0.995 & 1.000 & 0.997 & 0.999 & 1.000 & 1.000 & 1.000 & 1.000 & 0.995 & 0.995 & 0.995 & 0.998 \\
    SWEET & 
    0.985 & 1.000 & 0.992 & 0.998 & 1.000 & 0.995 & 0.997 & 0.999 & 0.990 & 1.000 & 0.994 & 0.999 & 0.980 & 1.000 & 0.990 & 0.990 \\
    Unigram & 
    0.995 & 1.000 & 0.997 & 0.998 & 0.995 & 1.000 & 0.997 & 0.999 & 1.000 & 1.000 & 1.000 & 1.000 & 0.990 & 1.000 & 0.994 & 0.999 \\
    Unbiased & 
    0.980 & 0.990 & 0.984 & 0.995 & 0.975 & 1.000 & 0.987 & 0.998 & 1.000 & 0.980 & 0.990 & 0.999 & 0.975 & 1.000 & 0.987 & 0.991 \\
    \midrule

    \rowcolor{sampling!30}\multicolumn{17}{c}{ {\textbf{Sampling Watermark}}} \\
    AAR & 
    0.995 & 1.000 & 0.997 & 0.999 & 0.995 & 1.000 & 0.997 & 0.995 & 1.000 & 1.000 & 1.000 & 1.000 & 0.995 & 1.000 & 0.997 & 0.999 \\
    EXP & 
    0.975 & 0.925 & 0.951 & 0.960 & 0.975 & 0.945 & 0.960 & 0.970 & 0.980 & 0.925 & 0.953 & 0.960 & 0.990 & 0.965 & 0.977 & 0.977 \\
    ITS & 
    0.965 & 0.950 & 0.957 & 0.968 & 0.980 & 0.985 & 0.982 & 0.987 & 0.925 & 0.890 & 0.909 & 0.928 & 0.985 & 0.970 & 0.978 & 0.979 \\
    GumbelSoft & 
    0.975 & 1.000 & 0.987 & 0.983 & 0.990 & 1.000 & 0.994 & 0.995 & 1.000 & 1.000 & 1.000 & 1.000 & 0.985 & 1.000 & 0.992 & 0.994 \\
    SynthID & 
    0.985 & 0.995 & 0.989 & 0.998 & 1.000 & 1.000 & 1.000 & 1.000 & 0.995 & 1.000 & 0.997 & 0.999 & 0.955 & 0.995 & 0.974 & 0.995 \\
    \midrule

    \rowcolor{symbiotic!30}\multicolumn{17}{c}{
    {\textbf{Symbiotic Watermark (Ours)}}} \\
    \textbf{Series} & \textbf{1.000} & \textbf{1.000} & \textbf{1.000} & \textbf{1.000} & \textbf{1.000} & \textbf{1.000} & \textbf{1.000} & \textbf{1.000} & \textbf{1.000} & \textbf{1.000} & \textbf{1.000} & \textbf{1.000} & \textbf{1.000} & \textbf{1.000} & \textbf{1.000} & \textbf{1.000} \\
    \textbf{Parallel} & 0.995 & 0.995 & 0.995 & 0.997 & 1.000 & 0.990 & 0.995 & 0.998 & 1.000 & 0.990 & 0.995 & 0.999 & 1.000 & 0.995 & 0.997 & 0.997\\ 
    \textbf{Hybrid} & 
    \textbf{0.995} & \textbf{1.000} & \textbf{0.997} & \textbf{0.999} & \textbf{1.000} & \textbf{1.000} & \textbf{1.000} & \textbf{1.000} & \textbf{1.000} & \textbf{1.000} & \textbf{1.000} & \textbf{1.000} & \textbf{0.995} & \textbf{1.000} & \textbf{0.997} & \textbf{0.999} \\
    
    \bottomrule[1.5pt]
    \end{tabular}}
    \caption{Detectability of OPT-6.7B and GPT-J-6B under different watermarking algorithms on C4 and OpenGen.}
    \label{table1}
\end{table*}

\subsection{Detectability}
Table ~\ref{table1} presents the overall watermark detection results for two datasets and four base models. 

\textit{Series scheme achieves state-of-the-art (SOTA) detectability performance.} 
Series scheme exhibits a perfect TPR of \textbf{1.000} across all datasets and models, signifying no false positives, which is crucial given the higher impact of false positives in watermarking contexts. This is due to the injection of double watermark signals into each token, reinforcing the watermark presence throughout the sequence. However, this enhanced detectability comes at the cost of text quality, as strong constraints are imposed on token selection at both the logits and sampling stages. 


\textit{Parallel scheme demonstrates competitive detectability performance}, with an average F1/AUC improvement of 1.60\%/1.35\% over sampling watermark. Despite each token being modified by only one of the two watermarking strategies (logits or sampling), sufficient watermark signal remains for detection. This result highlights that doubling watermarking is not always necessary for detection.

\textit{Hybrid scheme consistently outperforms baselines across various datasets and base model configurations, evidencing its remarkable generalization.} Specifically, Compared to the sampling watermark, Hybrid's F1/AUC performance improves by 1.90\%/1.52\% on average. This scheme adaptively assigns watermarking strategies based on entropy characteristics, which enables optimal watermark placement, ensuring high detectability while preserving text quality.

\begin{table*}[tp]
    \resizebox{\linewidth}{!}{
    \begin{tabular}{lcccccccccccccccc}
        \toprule[1.5pt]
        
        \multirow{2}{*}{\Large Model}&  \multicolumn{4}{c}{\textbf{T1: Short Q, Short A}} & \multicolumn{4}{c}{\textbf{T2: Short Q, Long A}} & \multicolumn{4}{c}{\textbf{T3: Long Q, Short A}} & \multicolumn{4}{c}{\textbf{T4: Long Q, Long A}} \\
        
        & \multicolumn{4}{c}{{\textit{Factual Knowledge}}} & \multicolumn{4}{c}{\textit{Long-form QA}} & \multicolumn{4}{c}{\textit{Reasoning \& Coding}} & \multicolumn{4}{c}{\textit{Summarization}} \\

        \cmidrule(lr){2-5}
        \cmidrule(lr){6-9}
        \cmidrule(lr){10-13}
        \cmidrule(lr){14-17}
                
        \large{+ Watermark} & TPR & TNR & GM & DROP & TPR & TNR & GM & DROP & TPR & TNR & GM & DROP & TPR & TNR & GM & DROP \\
        
        \hline \specialrule{0em}{2pt}{0pt}
        
        \makecell[l]{\large\textsc{LLaMA3-8B}} & - & - & 57.50 & - & - & - & 24.05 & - & - & - & 48.43 & - & - & - & 27.18 & - \\

        \specialrule{0em}{1pt}{1pt}
        
        + KGW & 0.815 & 0.700 & 56.00 & $\downarrow$ 2.61\% & 0.990 & 0.975 & 23.32 & $\downarrow$ 3.04\% & 0.740 & 0.845 & 36.40 & $\downarrow$ 24.8\% & 0.955 & 0.985 & 26.66 & $\downarrow$ 1.91\% \\

        + Unigram & 0.955 & 0.360 & 51.00 & $\downarrow$ 11.3\% & 0.965 & 0.990 & 23.24 & $\downarrow$ 3.37\% & 0.775 & 0.695 & 40.95 & $\downarrow$ 15.4\% & 0.915 & 0.890 & 26.89 & $\downarrow$ 1.07\% \\

        + EWD & 0.860 & 0.745 & 49.00 & $\downarrow$ 14.8\% & 1.000 & 1.000 & 23.52 & $\downarrow$ 2.20\% & 0.740 & 0.850 & 35.45 & $\downarrow$ 26.8\% & 0.965 & 0.990 & 26.68 & $\downarrow$ 1.84\% \\

        + AAR & 0.685 & 0.930 & 46.00 & $\downarrow$ 18.3\% & 0.995 & 1.000 & 21.95 & $\downarrow$ 8.73\% & 0.910 & 0.990 & 38.95 & $\downarrow$ 19.6\% & 1.000 & 0.995 & 25.14 & $\downarrow$ 7.51\% \\

        + SynthID & 0.780 & 0.530 & 51.00 & $\downarrow$ 11.3\% & 0.990 & 0.970 & 23.60 & $\downarrow$ 1.87\% & 0.790 & 0.695 & 39.10 & $\downarrow$ 19.3\% & 0.955 & 0.935 & 26.83 & $\downarrow$ 1.29\% \\
                
        + \textbf{Series} & 0.970 & 0.935 & 55.00 & $\downarrow$ 4.35\% & 0.950 & 1.000 & 21.82 & $\downarrow$ 9.27\% & 0.770 & 0.995 & 41.26 & $\downarrow$ 14.8\% & 0.930 & 1.000 & 26.22 & $\downarrow$ 3.53\% \\

        + \textbf{Parallel} & 0.965 & 0.450 & 52.00 & $\downarrow$ 9.57\% & 0.730 & 0.970 & 22.35 & $\downarrow$ 7.07\% & 0.765 & 1.000 & 42.63 & $\downarrow$ 12.0\% & 0.910 & 0.940 & 26.76 & $\downarrow$ 1.55\% \\

        + \textbf{Hybrid} & 1.000 & 0.960 & 57.00 & $\downarrow$ \textbf{0.87\%} & 0.965 & 1.000 & 23.61 & $\downarrow$ \textbf{1.83\%} & 0.925 & 0.990 & 42.65 & $\downarrow$ \textbf{11.9\%} & 0.965 & 0.995 & 26.92 & $\downarrow$ \textbf{0.96\%} \\
        
        \bottomrule[1.5pt]
    \end{tabular}}
    \caption{The performance of various watermarking algorithms across four different downstream tasks using True Positive Rate (TPR), True Negative Rate (TNR), Generation Metric (GM), and Generation Quality Drop (Drop).}
    \label{table2}
\end{table*}

\begin{figure}[tp]
\centering
\includegraphics[width=\linewidth]{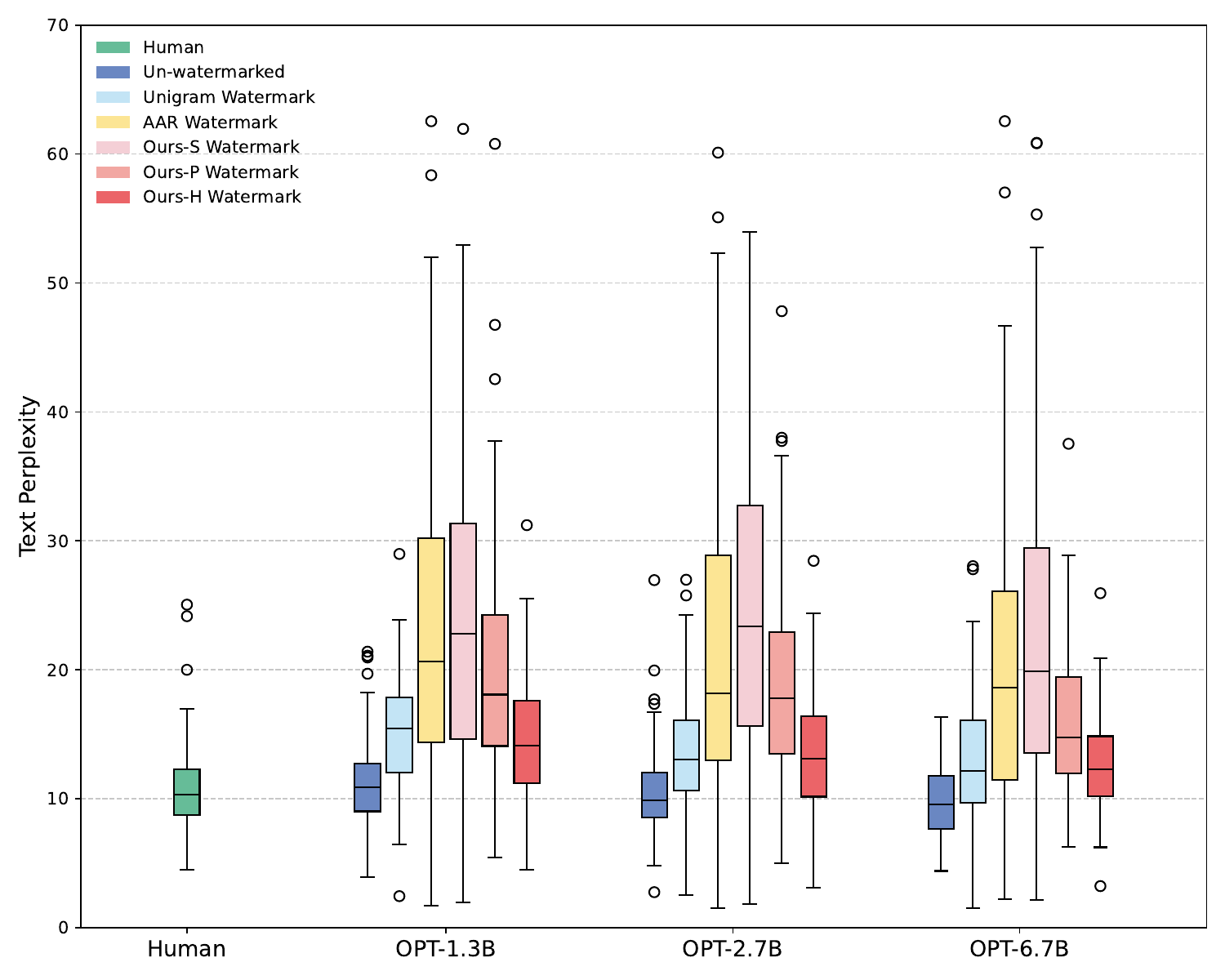}
\caption{A comparison of PPL across three symbiotic watermarking schemes with different model sizes.}
\label{figure 4}
\end{figure}

\subsection{Text Quality}
To evaluate the impact of our watermarking framework on text quality, we focus on perplexity and downstream tasks. Table \ref{table2} and Figure \ref{figure 4} show that our hybrid scheme achieves minimal performance drop and the lowest perplexity than baselines.

\paragraph{Perplexity.} To assess the fluency of watermarked text, we used LLaMA2-7B to compute the perplexity (PPL) of texts generated by models of varying sizes with different watermarking algorithms. As shown in Figure \ref{figure 4}, the Parallel scheme results in a lower PPL compared to the Serial scheme, as double watermarking per token degrades text quality more than single watermarking. Unlike Parallel watermarking, which groups tokens by odd and even positions, hybrid watermarking introduces semantic entropy and adaptively applies stage-specific watermarks, effectively managing text quality and achieving the lowest PPL.

\paragraph{Downstream Task.}
Fidelity is the cornerstone of watermarking algorithms, to further validate the impact of watermarking on text quality, we followed Waterbench \cite{tu-etal-2024-waterbench} settings and considered four downstream tasks (Details refer to Appendix \ref{Downstream Task Datasets}). The results in Table \ref{table2} indicate that the longer the generated answers (e.g., Task 2 and Task 4), the smaller the impact of the injected watermark on downstream tasks. Across all tasks, our hybrid scheme consistently achieves a high detection rate and superior task performance. Specifically, performance drops by only 0.87\% on Task 1 and 0.96\% on Task 4, demonstrating minimal distortion. 
Compared to baselines, SynthID imposes relatively minor text quality degradation but suffers from a lower detection rate, whereas other baselines exhibit either excessive text degradation or weaker detectability. In contrast, the Hybrid scheme strategically ensures strong detectability while preserving text fidelity, more suitable for real-world deployment.

\subsection{Robustness to Real-world Attacks}

\begin{figure*}[tp]
\centering
\includegraphics[width=\textwidth]{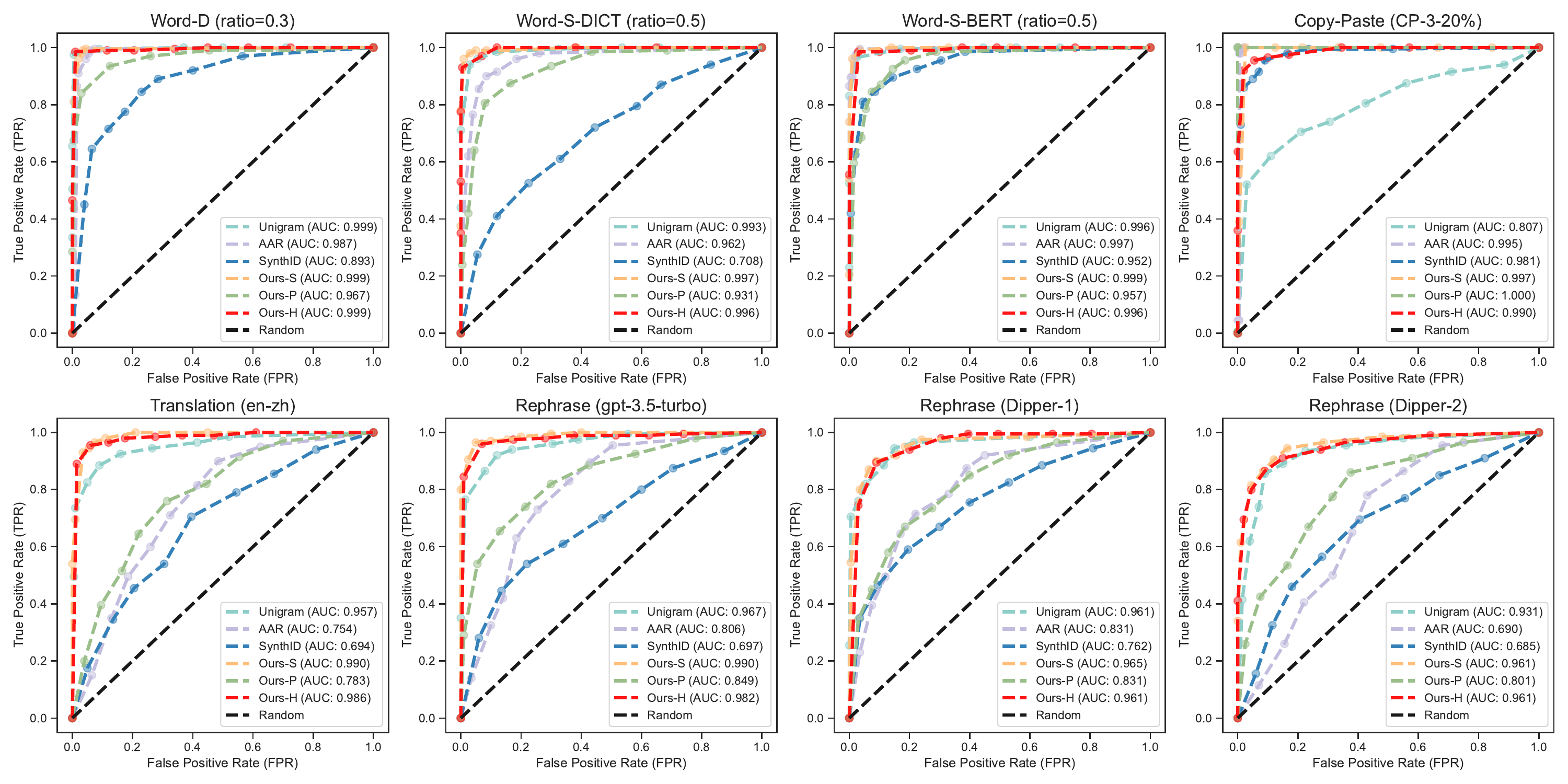}
\caption{The AUROC curve of watermarked text generated by OPT-6.7B under various attacks on C4 dataset.}
\label{figure 5}
\end{figure*}

Ensuring the robustness of watermarking schemes against various attacks is crucial for real-world applicability \cite{kirchenbauer2024on}. To provide comprehensive evidence of SymMark’s robustness, we conduct experiments to test its resilience against four attacks:\textbf{ Editing, Copy-Paste, Back-Translation, and Rephrasing}. Details are in Appendix \ref{Attack Settings}.



The ROC curves and AUC values for comparison in Figure \ref{figure 5} indicate \textbf{Hybrid's consistently robust watermark detection capabilities facing all attack scenarios}.
The average AUC values of serial and hybrid symbiotic watermarks are 0.987 and 0.984, respectively, significantly outperforming Unigram, the previously most robust method, with an AUC of 0.951. The Parallel scheme shows a relatively lower AUC, suggesting that injecting only one watermark signal per token is more vulnerable to adversarial modifications. 

\textbf{Hybrid excels in robustness is due to:}
(1) Dual-signal Injection. Hybrid ensures that even if one watermarking signal is partially disrupted, the other remains intact, enabling reliable detection;
(2) Entropy-driven Adaptation. Unlike fixed strategies, Hybrid is driven by entropy to adaptively select watermarking constraints, ensuring both imperceptibility and resilience;
(3) Cross-attack Generalization. While some methods perform well on specific attacks, Hybrid maintains high detection rates across diverse attack categories, making it practical for real-world deployment where adversarial conditions are unpredictable.

\subsection{Security}
\begin{figure}[tp]
\centering
\includegraphics[width=\linewidth]{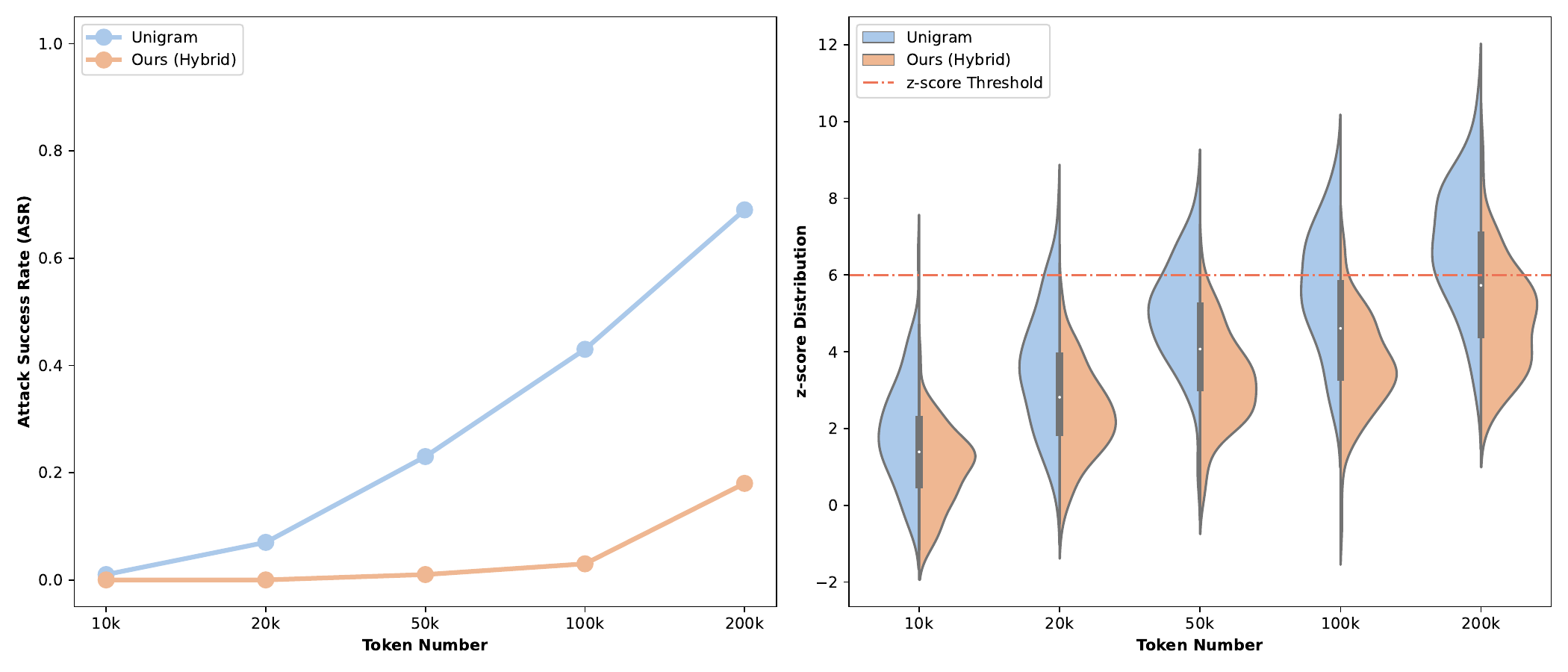}
\caption{The ASR of watermark stealing for varying numbers of tokens (left) and the z-score distribution of spoofing watermark (right) on LLaMA2-7B-chat-hf.}
\label{figure 6}
\end{figure}

Existing watermark stealing strategies, such as those targeting logits-based methods (e.g., the KGW family), are ineffective against sampling-based watermarks, which remain immune to such attacks. To explore the security of symbiotic watermarks, we apply the watermark stealing method and perform a spoofing attack \cite{jovanovic2024watermark, pang2024no} on the Unigram and our Hybrid. Detailed settings are in Appendix \ref{Watermark Stealing Settings}.

Figure \ref{figure 6} presents stealing results. The left panel depicts the Attack Success Rate (ASR) of watermark stealing, while the right panel presents the z-score distribution of spoofed Unigram and our Hybrid across different token counts. As the number of tokens obtained by the attacker increases, so does the ASR and z-score. However, the ASR and z-score of Hybrid scheme is much lower than that of the naive Unigram. When generating 200,000 tokens, the ASR for the original Unigram reaches 69\%, whereas the ASR for our symbiotic watermark scheme is only 18\%. 

The enhanced security of the Hybrid scheme stems from its non-linear combination of logits-based and sampling-based watermarking methods. Since the symbiotic watermarking rules are influenced not only by the logits but also by the inherent randomness in the sampling process, attackers are unable to reconstruct the watermarking rules purely through token frequency statistics or distribution modeling. This makes the Hybrid scheme significantly more resistant to watermark stealing attacks, offering enhanced security, particularly in adversarial environments where attackers are actively attempting to subvert the watermark.

\section{Conclusion}
This paper introduces a versatile symbiotic watermarking framework including three strategies: Serial, Parallel, and Hybrid. The Hybrid symbiotic watermark strategy leverages token and semantic entropy to balance detectability, robustness, text quality, and security. Experimental results across various datasets and models demonstrate the effectiveness of our method, shifting the focus from trade-offs to synergy. In the future, we will explore additional symbiotic watermarking paradigms, investigating perspectives beyond entropy to further advance watermarking techniques.

\section{Limitations}

This paper explores combining logits-based and sampling-based watermarks from an entropy perspective, while acknowledging that entropy is not the only evaluation metric. Future research could adopt other mathematical or information-theoretic tools to enhance symbiotic watermark design. Metrics like information gain and signal-to-noise ratio, alongside entropy, may offer deeper insights into watermark performance, robustness, and efficiency. These metrics can support the development of more adaptable watermarking schemes for diverse applications. Considering alternative metrics may lead to more flexible watermark designs suitable for varied scenarios. Despite limitations, we believe the symbiotic watermark concept offers a novel perspective and meaningful direction for advancing LLM watermarking in this fast-evolving field.

\section{Ethical Statement}
With the rapid development of large language models (LLMs) and their widespread applications, incorporating watermarks into LLM-generated content facilitates traceability, thereby significantly enhancing transparency and accountability. Building on previous research, this paper seeks to achieve a balance among the detectability, text quality, security, and robustness of watermarks. We aspire for the framework proposed in this paper to offer novel insights into watermarking methodologies and to be further utilized in safeguarding intellectual property, curbing misinformation, and mitigating AIGC misuse, including academic fraud, thereby fostering public trust in AI technologies.


\bibliography{main, anthology}

\appendix

\section{Efficient Analysis}
\begin{table}[htbp]
    \centering
    \resizebox{\linewidth}{!}{\begin{tabular}{c|cccccc}
     \toprule
     \toprule
     Method & KGW & AAR & EXP & Series & Parallel & Hybrid \\
     \midrule
     Generation Time & 8.475s & 8.605s & 8.260s & 8.745s & 12.675s & 15.575s \\
     Detection Time & 0.035s & 0.045s & 65.74s & 0.050s & 0.060s & 0.050s \\
     \bottomrule
     \bottomrule
    \end{tabular}}
    \caption{The computational efficiency analysis of different watermarking for each text of length 200 tokens.}
    \label{efficient analysis}
\end{table}

All experiments were conducted on two NVIDIA A100 GPUs. Table \ref{efficient analysis} presents the average time required by several representative watermarking methods to generate and detect watermark texts of 200 tokens using OPT-6.7B. Our symbiotic watermarking strategy achieves nearly the same efficiency as existing methods in watermark detection. Although our hybrid watermarking method incurs additional computation time for token and semantic entropy during watermark text generation, this overhead remains acceptable in practical applications and contributes to enhanced robustness, security, and text quality. Furthermore, this overhead could be mitigated if entropy calculation were integrated into the Hugging Face\footnote{https://huggingface.co/} tool library in the future.

\section{Distinguishing Human-Written Text}\label{Distinguishing Human-Written Text}

\begin{figure}[htbp]
\centering
\includegraphics[width=\linewidth]{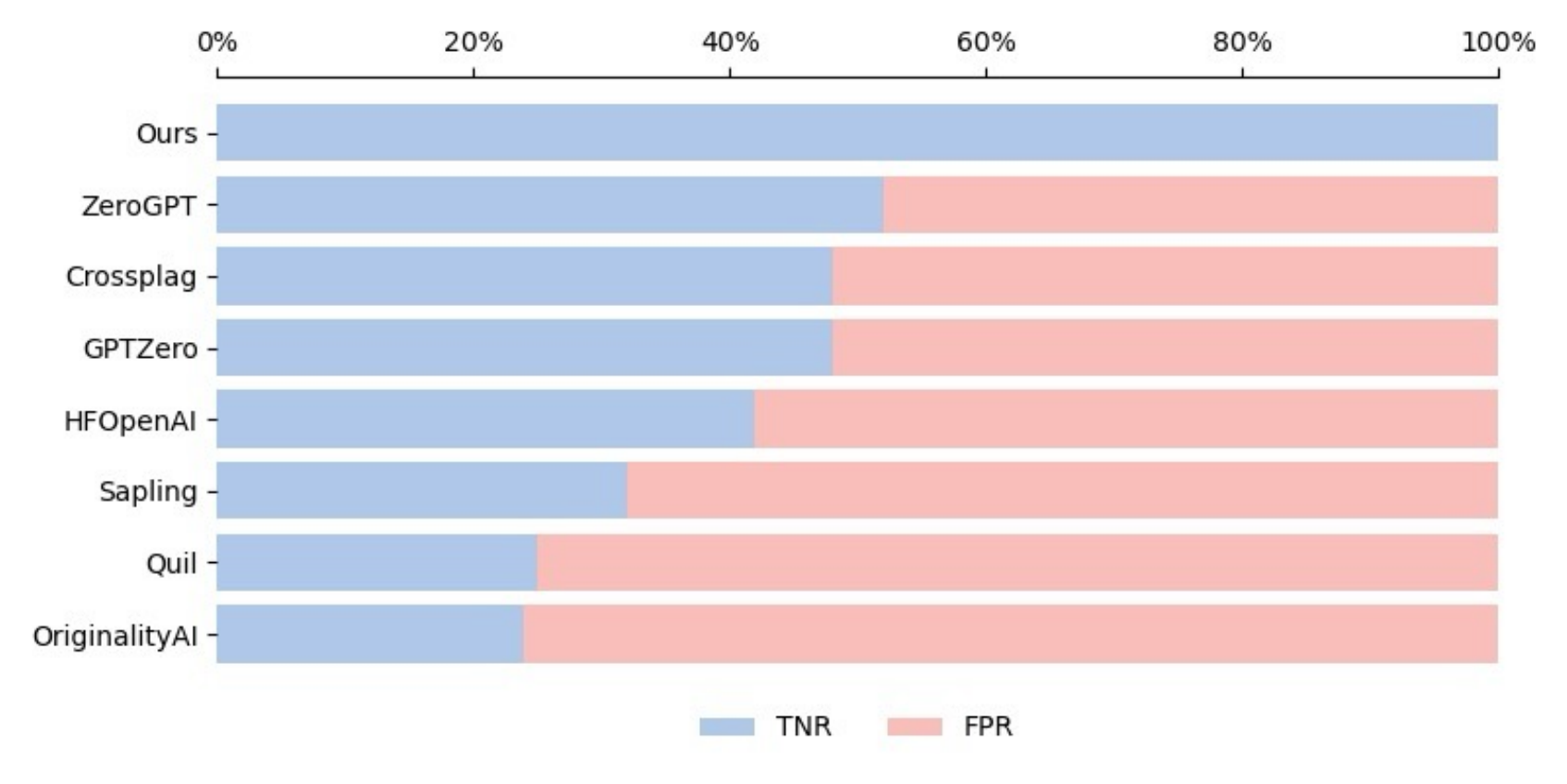}
\caption{Comparing AIGT detection methods and ours in distinguishing human-written text on TOEFL dataset.}
\label{figure 7}
\end{figure}

Based on \citet{liang2023gpt}, we evaluated our method using the TOEFL dataset, comprising non-native English writing samples, as shown in Figure \ref{figure 7}. The experimental results show that our approach reliably identifies text with watermarks while non-native English writing samples are susceptible to misclassification by existing AIGT (AI-generated text) detection methods. These findings highlight the practicality and reliability of our watermarking method, which achieves a near-zero FPR.

\section{Downstream Task Datasets}\label{Downstream Task Datasets}
Referring to Waterbench \cite{tu-etal-2024-waterbench}, we utilize the following datasets: 

\begin{itemize}[leftmargin=*, topsep=0pt]
    \item \textbf{Category 1 (Short Input, Short Answer)} includes the concept-probing Copen dataset \cite{peng-etal-2022-copen}, with 200 samples selected from the CIC and CSJ tasks. Given the short output length, the \textbf{F1 score} is chosen as the evaluation metric. The max\_new\_tokens parameter for model generation is set to \textbf{16}.
    \item \textbf{Category 2 (Short Input, Long Answer)} utilizes 200 samples from the ELI5 dataset \cite{fan-etal-2019-eli5}, a long-form question-answering dataset originating from the Reddit forum “Explain Like I’m Five.” \textbf{Rouge-L} is employed as the evaluation metric. The max\_new\_tokens parameter for model generation is set to \textbf{300}.
    \item \textbf{Category 3 (Long Input, Short Answer)} addresses the code completion task, utilizing 200 samples from the LCC dataset \cite{chen2021evaluating}. This dataset is created by filtering single-file code samples from GitHub, with the \textbf{Edit Similarity} metric adopted for evaluation. The max\_new\_tokens parameter for model generation is set to \textbf{64}.
    \item \textbf{Category 4 (Long Input, Long Answer)} involves 200 samples from the widely-used MultiNews dataset \cite{fabbri-etal-2019-multi}, a multi-document news summarization dataset.     \textbf{Rouge-L} serves as the evaluation metric. The max\_new\_tokens parameter for model generation is set to \textbf{512}.
\end{itemize}

\section{Baseline Settings}\label{Baseline Settings}
\begin{figure*}[htbp]
\centering
\includegraphics[width=\textwidth]{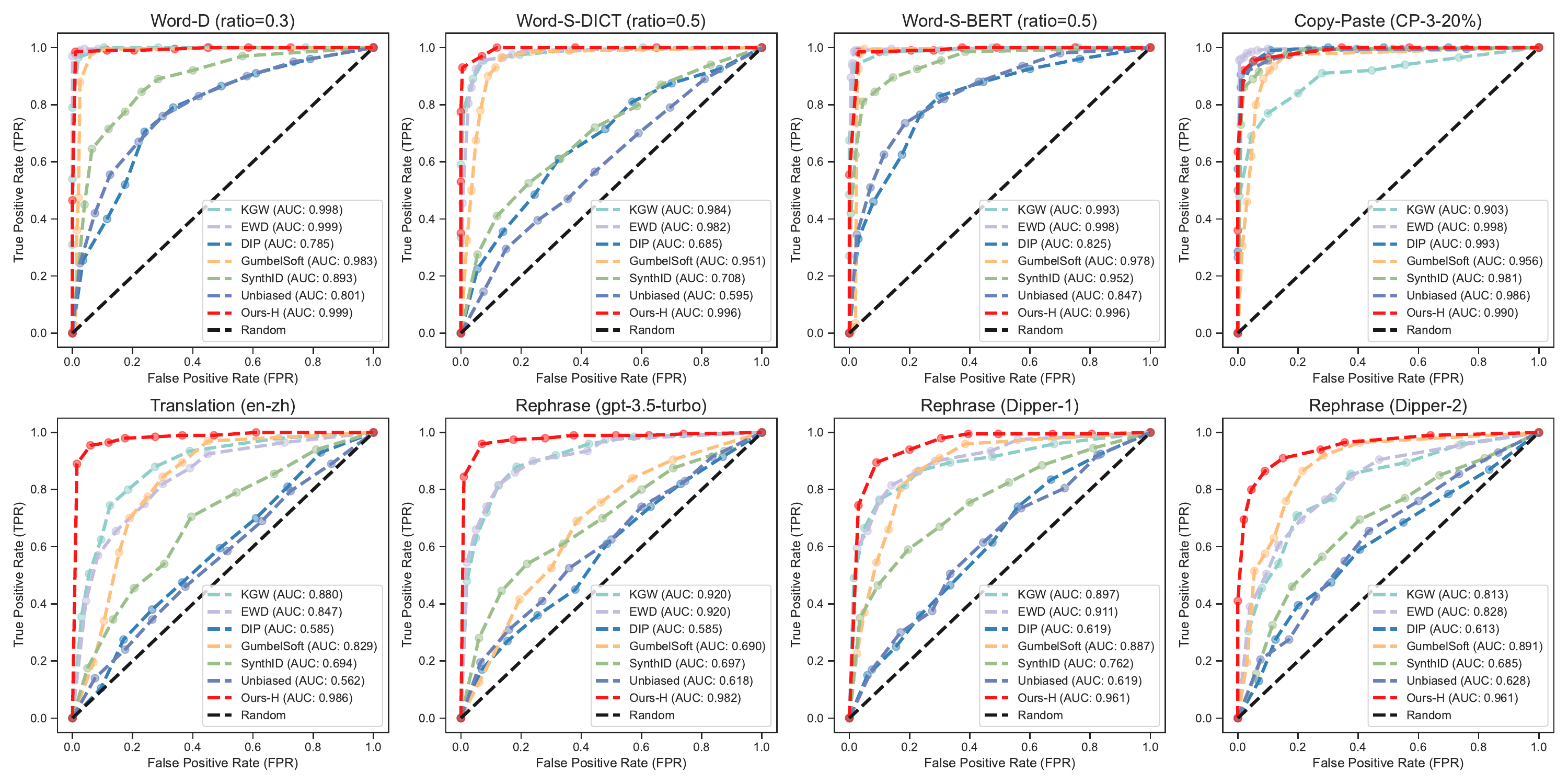}
\caption{The AUROC curve of watermarked text generated by OPT-6.7B under various attacks on C4 dataset.}
\label{figure 8}
\end{figure*}

We use MarkLLM \cite{pan-etal-2024-markllm} toolkit to implement both the baseline and our proposed method, as detailed below:

\begin{itemize}[leftmargin=*]
    \item \textbf{KGW} proposed by \citet{pmlr-v202-kirchenbauer23a}, the details of the parameters are as follows: $\gamma$ = 0.5, $\delta$ = 0.2, $\xi$ = 15485863, prefix\_length = 1, z\_threshold = 4.0, window\_scheme = "left".

    \item \textbf{Unigram} proposed by \citet{zhao2024provable}, the details of the parameters are as follows: $\gamma$ = 0.5, $\delta$ = 2.0, $\xi$ = 15485863, z\_threshold = 4.0

    \item \textbf{DIP} proposed by \citet{wu2024a}, the details of the parameters are as follows: $\gamma$ = 0.5, $\alpha$ = 0.45, key = 42,prefix\_length = 5, z\_threshold=1.513

    \item \textbf{SWEET} proposed by \citet{lee-etal-2024-wrote}, the details of the parameters are as follows: $\gamma$ = 0.5, $\delta$ = 2.0, $\xi$ = 15485863, prefix\_length = 1, z\_threshold = 4.0, entropy\_threshold = 0.9

    \item \textbf{EWD} proposed by \citet{lu-etal-2024-entropy}, the details of the parameters are as follows: $\gamma$ = 0.5, $\delta$ = 2.0, $\xi$ = 15485863, prefix\_length = 1, z\_threshold=4.0

    \item \textbf{Unbiased} proposed by \citet{hu2024unbiased}, the details of the parameters are as follows: $\gamma$ = 0.5, key = 42, prefix\_length = 5, z\_threshold=1.513

    \item \textbf{AAR} proposed by \citet{Aaronson}, the details of the parameters are as follows: prefix\_length = 4, $\xi$ = 15485863, p\_value = 1e-4, sequence\_length = 200

    \item \textbf{EXP} proposed by \citet{kuditipudi2024robust}, the details of the parameters are as follows: pseudo\_length = 420, sequence\_length = 200, n\_runs = 100, key = 42, p\_threshold = 0.2

    \item \textbf{ITS} proposed by \citet{kuditipudi2024robust}, the details of the parameters are as follows: pseudo\_length = 256, sequence\_length = 200, n\_runs = 500, key = 42,
    p\_threshold = 0.1

    \item \textbf{GumbelSoft} proposed by \citet{fu-etal-2024-gumbelsoft}, the details of the parameters are as follows: prefix\_length = 2, eps = 1e-20, threshold = 1e-4, sequence\_length = 200, temperature = 0.7

    \item \textbf{SynthID} proposed by \citet{dathathri2024scalable}, the details of the parameters are as follows: n = 5, sampling\_size = 65536, seed = 0, mode = "non-distortionary", num\_leaves = 2, context\_size = 1024, detector\_type = "mean", threshold = 0.52
\end{itemize}

\section{Watermark Selection}

In our symbiotic framework SymMark, we adopt the Unigram method \cite{zhao2024provable} for logits-based watermarking, as it surpasses the KGW algorithm \cite{pmlr-v202-kirchenbauer23a} in robustness and maintains relatively high text quality compared to other logits-based watermarking methods, including Unbiased, DIP, and SWEET. For sampling-based watermarking, we select the AAR \cite{Aaronson} algorithm to improve both robustness and security. This choice is motivated by the extremely low detection efficiency of the EXP and ITS \cite{kuditipudi2023robust} watermarks, as shown in Table \ref{efficient analysis}, along with the relatively poor detectability of both GumbelSoft \cite{fu-etal-2024-gumbelsoft} and SynthID \cite{dathathri2024scalable}. The parameter settings remain identical to the baselines.

\begin{figure}[tp]
\centering
\includegraphics[width=\linewidth]{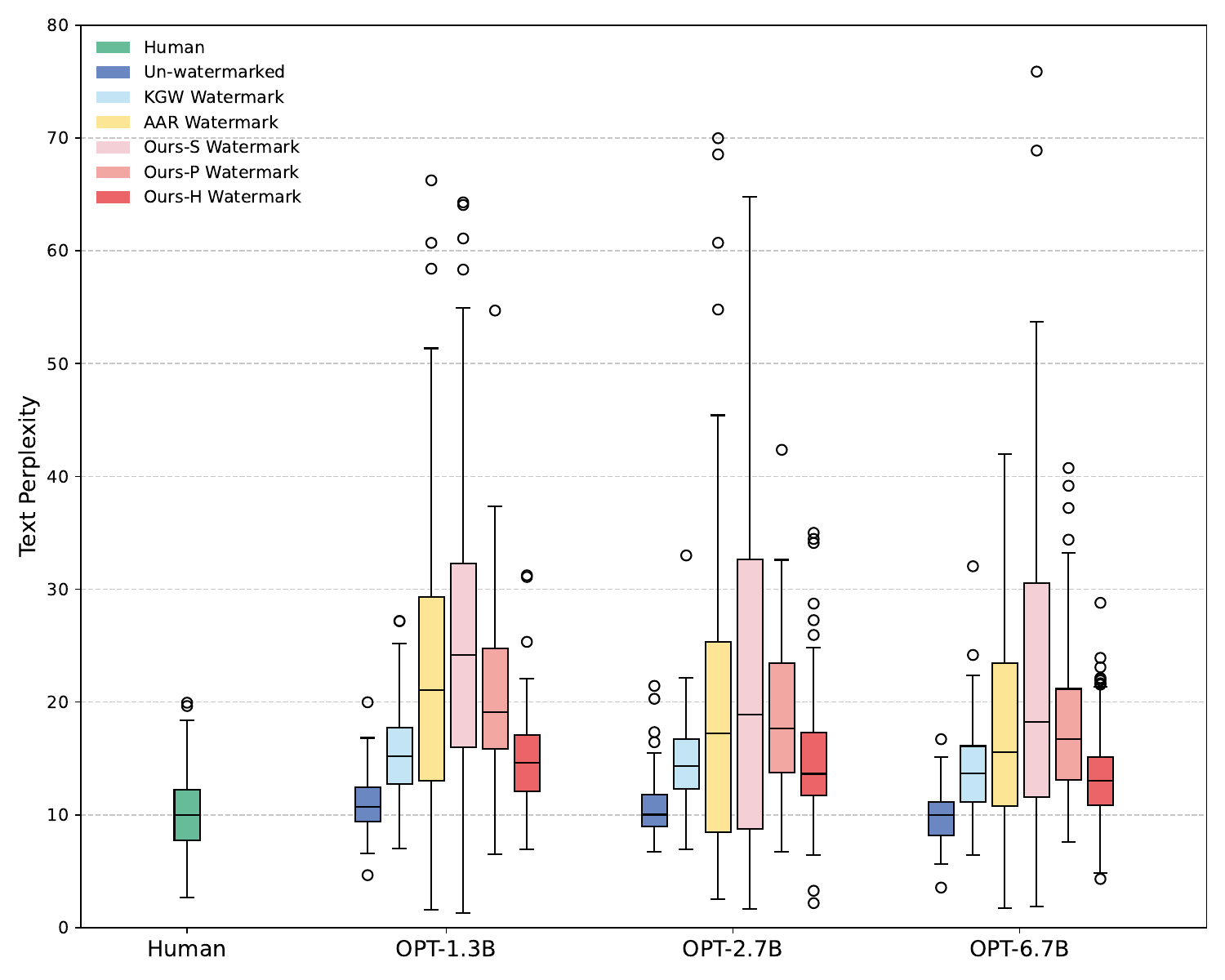}
\caption{A comparison of PPL across three symbiotic watermarking schemes with different model sizes.}
\label{figure 9}
\end{figure}

\begin{table*}[tp]
    \centering
    \resizebox{\textwidth}{!}{\begin{tabular}{ccccccccccccccccc}
    \toprule[1.5pt]
    \multirow{3}{*}{Watermark} & \multicolumn{8}{c}{\textsc{C4 Dataset}} & \multicolumn{8}{c}{\textsc{OpenGen Dataset}}\\

    & \multicolumn{4}{c}{\textsc{OPT-6.7B}} & \multicolumn{4}{c}{\textsc{GPT-J-6B}} & \multicolumn{4}{c}{\textsc{OPT-6.7B}} & \multicolumn{4}{c}{\textsc{GPT-J-6B}}\\
    
    \cmidrule(lr){2-5}
    \cmidrule(lr){6-9} \cmidrule(lr){10-13}
    \cmidrule(lr){14-17}
    
    & TPR & TNR & F1 & AUC & TPR & TNR & F1 & AUC & TPR & TNR & F1 & AUC & TPR & TNR & F1 & AUC \\

    \midrule
    
    \rowcolor{symbiotic!30}\multicolumn{17}{c}{{ \textbf{KGW + AAR Watermark}}} \\
    Series & 1.000 & 0.995 & 0.998 & 0.999 & 1.000 & 1.000 & 1.000 & 1.000 & 1.000 & 0.995 & 0.998 & 0.999 & 1.000 & 1.000 & 1.000 & 1.000 \\
    Parallel & 1.000 & 0.970 & 0.985 & 0.990 & 1.000 & 0.980 & 0.990 & 0.992 & 0.995 & 0.955 & 0.975 & 0.976 & 0.985 & 0.980 & 0.983 & 0.985 \\
    Hybrid & 0.995 & 1.000 & 0.997 & 0.999 & 1.000 & 1.000 & 1.000 & 1.000 & 0.995 & 1.000 & 0.998 & 0.999 & 0.995 & 0.995 & 0.995 & 0.997 \\
    \midrule

    \rowcolor{symbiotic!30}\multicolumn{17}{c}{{\textbf{Unbiased + AAR Watermark}}} \\
    Series & 0.985 & 1.000 & 0.993 & 0.999 & 1.000 & 1.000 & 1.000 & 1.000 & 1.000 & 1.000 & 1.000 & 1.000 & 0.995 & 1.000 & 0.997 & 0.997 \\
    Parallel & 0.835 & 1.000 & 0.918 & 0.914 & 0.890 & 1.000 & 0.942 & 0.954 & 0.885 & 0.990 & 0.934 & 0.957 & 0.945 & 1.000 & 0.972 & 0.974 \\
    Hybrid & 0.970 & 1.000 & 0.985 & 0.994 & 0.920 & 1.000 & 0.956 & 0.973 & 0.995 & 1.000 & 0.997 & 0.998 & 0.965 & 1.000 & 0.982 & 0.992 \\
    \midrule

    \rowcolor{symbiotic!30}\multicolumn{17}{c}{{\textbf{KGW + GumbelSoft Watermark}}} \\
    Series & 0.985 & 1.000 & 0.992 & 0.993 & 0.970 & 1.000 & 0.985 & 0.988 & 1.000 & 1.000 & 1.000 & 0.996 & 0.975 & 1.000 & 0.987 & 0.996 \\
    Parallel & 0.935 & 1.000 & 0.967 & 0.992 & 0.955 & 0.995 & 0.974 & 0.993 & 0.980 & 0.990 & 0.985 & 0.995 & 0.900 & 1.000 & 0.947 & 0.997 \\
    Hybrid & 0.955 & 1.000 & 0.977 & 0.998 & 0.985 & 1.000 & 0.992 & 0.994 & 0.980 & 0.995 & 0.987 & 0.999 & 0.950 & 0.990 & 0.969 & 0.993 \\
    \midrule

    \rowcolor{symbiotic!30}\multicolumn{17}{c}{{\textbf{Unigram + GumbelSoft Watermark}}} \\
    Series & 0.995 & 1.000 & 0.997 & 0.995 & 0.995 & 0.980 & 0.988 & 0.999 & 0.975 & 0.995 & 0.985 & 0.999 & 0.995 & 0.995 & 0.995 & 0.996 \\
    Parallel & 0.870 & 1.000 & 0.930 & 0.993 & 0.985 & 0.955 & 0.970 & 0.978 & 0.920 & 0.985 & 0.951 & 0.981 & 0.940 & 0.965 & 0.952 & 0.993 \\
    Hybrid & 0.955 & 1.000 & 0.977 & 0.994 & 0.960 & 0.975 & 0.967 & 0.999 & 0.980 & 1.000 & 0.990 & 0.999 & 0.990 & 0.990 & 0.990 & 0.995 \\
    
    \bottomrule[1.5pt]
    \end{tabular}}
    \caption{Evaluating the detectability of different symbiotic watermarking algorithms on C4 and OpenGen.}
    \label{other combinations}
\end{table*}

We explored additional watermark combinations, with detection results summarized in Table \ref{other combinations}. Theoretically, both the KGW family (Unigram, SWEET, etc.) and the ARR family (EXP, GumbelSoft, etc.) can be integrated into our framework. As shown in Figure \ref{figure 9}, the corresponding PPL results of KGW and AAR further validate that our hybrid symbiotic watermarking strategy effectively balances detectability and text quality.

\section{Attack Settings} \label{Attack Settings}

Besides the method presented in Figure \ref{figure 5}, the AUROC curves for the attack robustness tests of the other baseline methods are illustrated in Figure \ref{figure 8}. The specific parameter settings for various attack scenarios are as follows:

\begin{itemize}[leftmargin=*]
    \item \textbf{Word-D} Randomly delete 30\% of the words in the watermark text.
    \item \textbf{Word-S-DICT} Replace 50\% of the words with their synonyms based on the WordNet \cite{miller1995wordnet} dictionary.
    \item \textbf{Word-S-BERT} Replace 50\% of the words with contextually appropriate synonyms using BERT’s \cite{devlin2018bert} embeddings.
    \item \textbf{Copy-Paste} Only 20\% of the watermark text is retained, distributed across three locations in the document.
    \item \textbf{Translation} Translate the text from English to Chinese and then back to English using the fine-tuned T5 translation model \footnote{https://huggingface.co/utrobinmv/}.
    \item \textbf{Rephrase (GPT-3.5-turbo)} Call GPT-3.5-turbo API to paraphrase the text with low creativity (temperature = 0.2).
    \item \textbf{Rephrase (Dipper-1)} Use the DIPPER \cite{krishna2023paraphrasing} model for a restatement attack, focusing on lexical diversity without changing sentence structure. (lex\_diversity = 60, order\_diversity = 0, max\_new\_tokens = 200)
    \item \textbf{Rephrase (Dipper-2)} Use DIPPER again, with both lexical and order diversity, generating even more varied restatements. (lex\_diversity=60, order\_diversity=60, max\_new\_tokens=200)
    
\end{itemize}

\section{Hyperparameter Analysis} \label{Hyperparameter Analysis}

\begin{figure*}[tp]
\centering
\includegraphics[width=\textwidth]{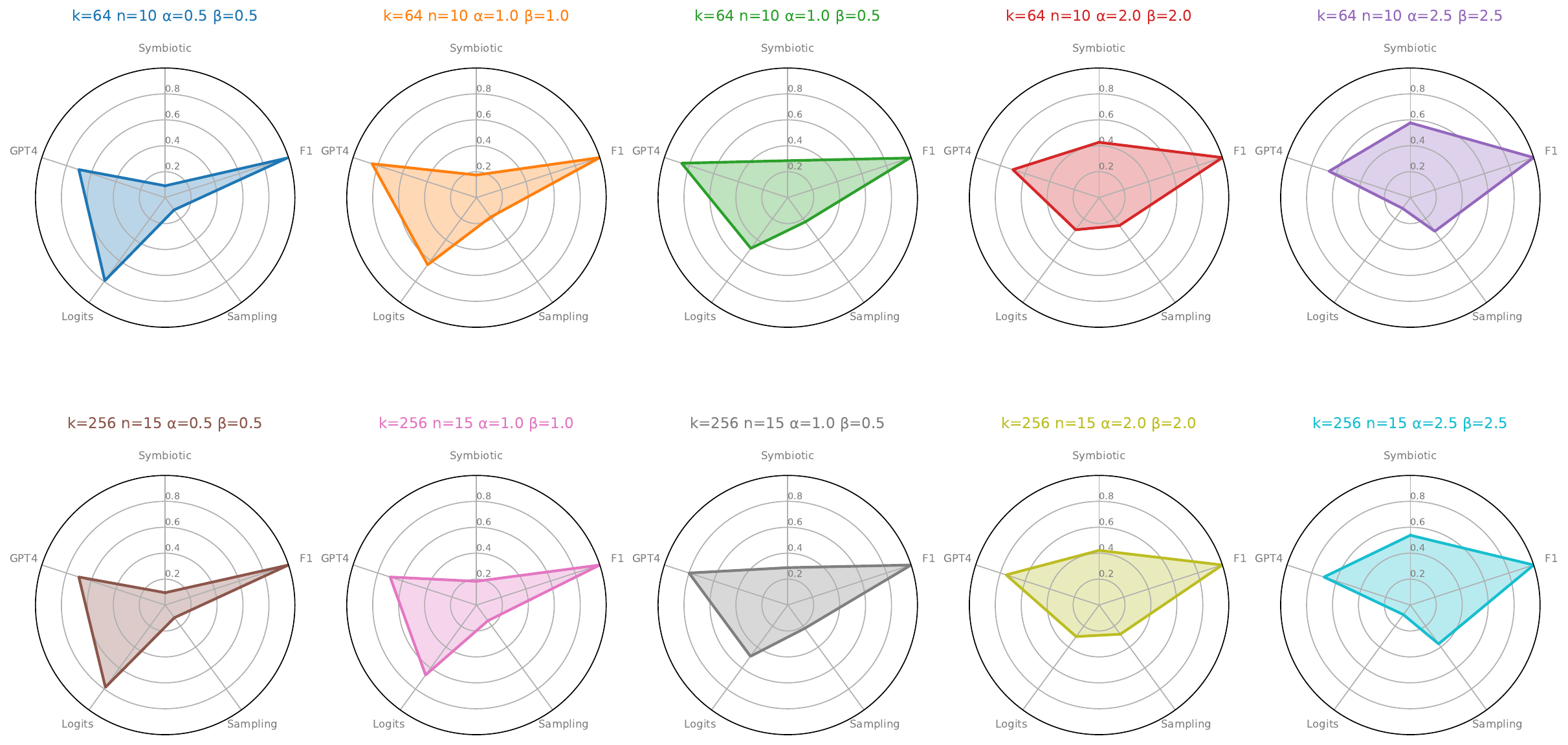}
\caption{Hyperparameter Analysis of Top-$k$ Selection, Number of Clusters $n$, TE threshold $\alpha$ and SE threshold $\beta$.}
\label{figure 10}
\end{figure*}

We randomly sampled 50 instances from the C4 dataset and embedded our hybrid symbiotic watermarks into the OPT-6.7B model. We analyzed the detection F1 scores and GPT-4’s evaluations of text quality under varying token entropy and semantic entropy thresholds, with the results displayed in Figure \ref{figure 10}. The prompt used for GPT-4 \cite{openai2023gpt} to evaluate watermarked text quality in Figure \ref{figure 10} and Figure \ref{figure 11} is as follows:

\tcbset{colframe = darkgray, fonttitle = \bfseries}
\begin{tcolorbox}[title = {GPT-4 Judge}] 
"You are given a prompt and a response, and you need to grade the response out of 100 based on: Accuracy (20 points) - correctness and relevance to the prompt; Detail (20 points) - comprehensiveness and depth; Grammar and Typing (30 points) - grammatical and typographical accuracy; Vocabulary (30 points) - appropriateness and richness. Deduct points for shortcomings in each category. Note that you only need to give an overall score, no explanation is required."
\end{tcolorbox}

\paragraph{The impact of top-$k$ and cluster number $n$.} As shown in Figure \ref{figure 10}, under different top-$k$ and $n$ settings, the variations in F1 and GPT-4 scores closely follow the changes in the entropy threshold. This indicates that top-$k$ and the number of clusters have minimal impact on semantic entropy calculation. Therefore, for clustering efficiency, we set top-$k$ to 64 and $n$ to 10.

\paragraph{The impact of entropy thresholds $\alpha$ and $\beta$.} In Figure \ref{figure 10}, “Symbiotic” represents the ratio of embedding logits to sampling watermarked tokens, “Logits” denotes the ratio of embedding logits watermark tokens, and “Sampling” refers to the ratio of embedding sampling watermark tokens. When the token and semantic entropy thresholds are low, the proportion of symbiotic watermarks remains low. As these thresholds increase, the proportion of symbiotic watermarks correspondingly rises. The two extreme cases of hybrid watermarks, corresponding to series and parallel configurations, constrain the impact of entropy thresholds on the detectability F1 score. However, an increased proportion of symbiotic watermarks more significantly affects text quality. Based on our experiments on the demo dataset, we set the token entropy threshold ($\alpha$) to 1.0 and semantic entropy threshold ($\beta$) to 0.5 to achieve an optimal trade-off between detectability and text quality.



\section{The impact of Semantic Entropy}
\label{The impact of Semantic Entropy}

\begin{figure*}[tp]
\centering
\includegraphics[width=\textwidth]{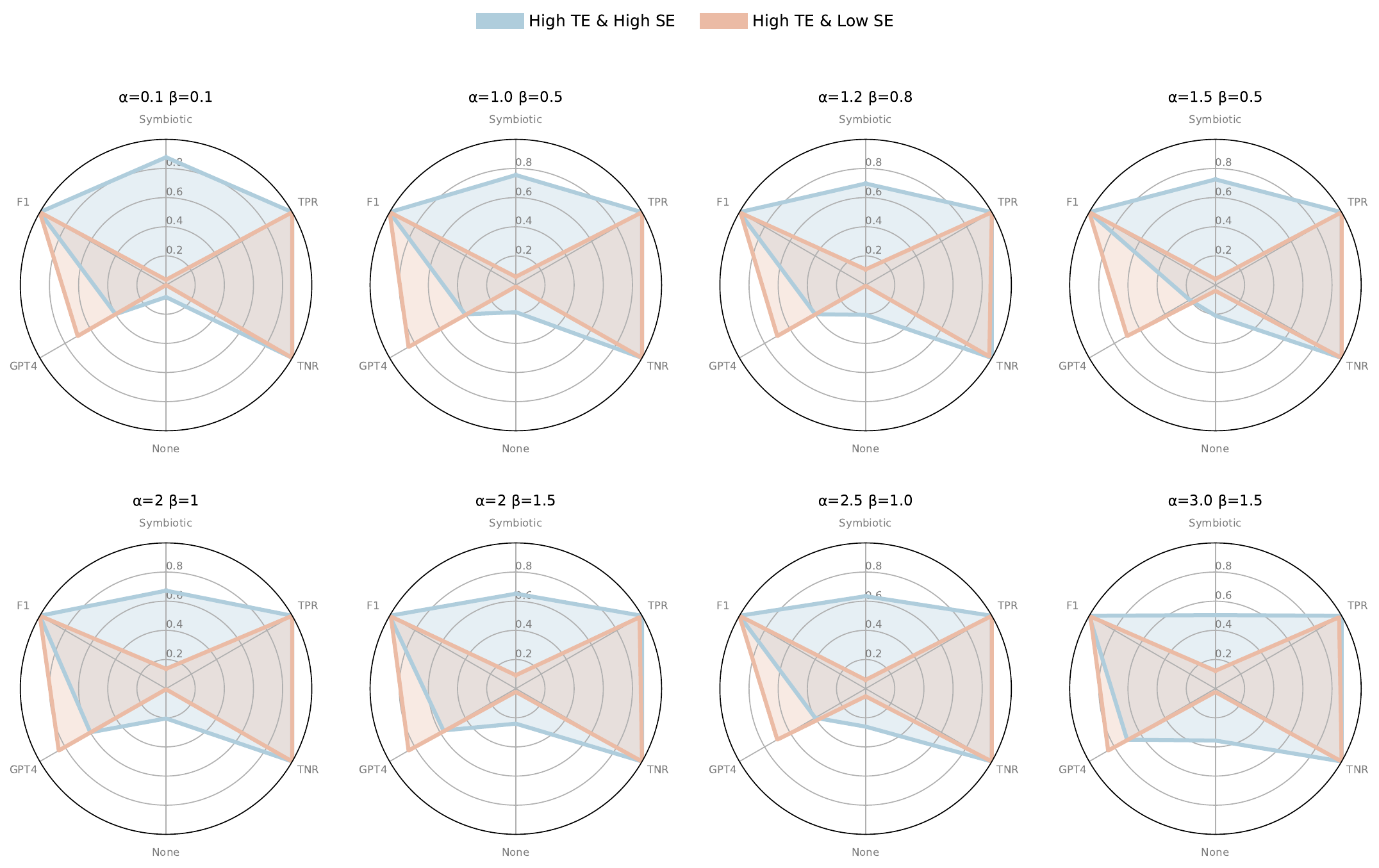}
\caption{Comparison of two watermarking schemes: high versus low token and semantic entropy. “Symbiotic” refers to embedding logits and sampling watermarked tokens, while “None” refers to unwatermarked tokens.}
\label{figure 11}
\end{figure*}

\begin{figure}[tp]
\includegraphics[width=\linewidth]{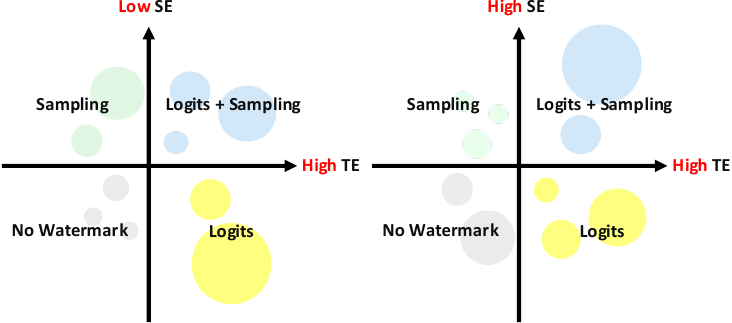}
\caption{Scheme 1 (Left), Scheme 2 (Right)}
\label{figure 12}
\end{figure}

We compared two entropy combination schemes:

\begin{itemize}[leftmargin=*]
    \item \textbf{Scheme 1} (we adopted): Embeds symbiotic watermarks at high token entropy and low semantic entropy.
    \item \textbf{Scheme 2}: Embeds symbiotic watermarks at high token entropy and high semantic entropy.
\end{itemize}

The experimental results for various token and semantic entropy thresholds are shown in Figure \ref{figure 11} and \ref{figure 12}. While both schemes demonstrate good detectability, Scheme 1 (GPT-4) significantly outperforms Scheme 2 in text quality assessment. This suggests that embedding watermarks on tokens with low semantic entropy has a lesser impact on text quality than embedding them on tokens with high semantic entropy. Even when watermarks are applied to tokens with low semantic entropy, the semantic integrity of the sampled tokens remains largely unchanged.

Furthermore, our experiments show that when token entropy is low, semantic entropy is also low, while when token entropy is high, semantic entropy can vary between high and low. Consequently, in many samples, numerous tokens are not embedded with the watermark in Scheme 2, negatively affecting watermark detection performance. In contrast, Scheme 1 successfully embeds sufficient watermark signals in nearly all cases, while preserving the text quality. Therefore, we choose to embed two watermark signals when token entropy is high and semantic entropy is low.

\section{Group-based Detection}\label{Group-based Detection}

We also considered a group-based detection algorithm, as shown in Algorithm \ref{Algorithm 3}. Specifically, we first group tokens into logits-based and sampling-based categories. In serial watermarks, each token contains two watermarks, so all tokens are grouped. For parallel watermarks, tokens are grouped by odd and even positions. In hybrid watermarks, we calculate the token and semantic entropy and group tokens based on entropy values. After grouping, we apply the logit-based and sampling-based watermark detection methods from Algorithm \ref{Algorithm 4}. However, this grouping approach has several drawbacks: (1) A more complicated detection process; (2) Low detection efficiency, especially for mixed symbiotic watermarks due to entropy calculations; (3) Poor robustness, as parallel watermarks' odd and even positions may change.


Therefore, this paper employs Algorithm \ref{Algorithm 2} for detection, as it directly identifies watermark signals in all tokens of the generated text. This method has demonstrated outstanding practical performance, is easy to implement, and ensures high watermark detection efficiency, as shown in Table \ref{efficient analysis}.	

\begin{algorithm}[tp]
\footnotesize
  \SetAlgoLined
  \KwIn{$\mathcal{M}, y_{1:T}, \alpha, \beta, \textsc{flag}$}
  \KwOut{$Y_l, Y_s$}
  \tcp{\textcolor{gray}{\mbox{Serial Watermark Group}}}
  \If{$\textsc{flag}$ = "S"}{
    $Y_l \leftarrow y_{1:T}$ \\
    $Y_s \leftarrow y_{1:T}$
  }
  \tcp{\textcolor{gray}{\mbox{Parallel Watermark Group}}}
  \ElseIf{$\textsc{flag}$ = "P"}{
    \If{$i\bmod2==0$}{$Y_l\text{.append}(y_i)$}
    \ElseIf{$i\bmod2==1$}{$Y_s\text{.append}(y_i)$}
  }
  \tcp{\textcolor{gray}{\mbox{Hybrid Watermark Group}}}
  \ElseIf{$\textsc{flag}$ = "H"}{
    \For{$i=1,...,T$}{
        \mbox{$H_{TE}, H_{SE} \leftarrow \text{ComputeEntropy}(y_{1:i})$}
        
        \tcp{\textcolor{gray}{\mbox{High Token Entropy}}}\If{$H_{TE}>\alpha$}{$Y_l\text{.append}(y_i)$}
        \tcp{\textcolor{gray}{\mbox{Low Semantic Entropy}}}
        \If{$H_{SE}<\beta$}{$Y_s\text{.append}(y_i)$}
    }
  }
  
 \caption{\mbox{Group Watermarked Token}}
 \label{Algorithm 3}
\end{algorithm}

\begin{algorithm}[tp]
\footnotesize
  \SetAlgoLined
  \KwIn{$\mathcal{M}, Y_{l}, Y_{s}, \mathcal{D}_{l}, \mathcal{D}_{s}, z_1, z_2$}
  \KwOut{$I\text{: True (Watermarked) or False}$}
  $I_l\leftarrow \text{False}$
  
  $I_s\leftarrow \text{False}$

  \tcp{\textcolor{gray}{\mbox{Logits Watermark Detection}}}\If{$\mathcal{D}_l(\mathcal{M},Y_l)>z_1$}{$I_l\leftarrow \text{True}$} 
  
  \tcp{\textcolor{gray}{\mbox{Sampling Watermark Detection}}}\If{$\mathcal{D}_s(\mathcal{M},Y_s)>z_2$}  {$I_s\leftarrow \text{True}$}

  \tcp{\textcolor{gray}{\mbox{Combine Detection Results}}}
  $I\leftarrow I_l \mid I_s$
 \caption{\mbox{Group-based Detection}}
 \label{Algorithm 4}
\end{algorithm}

\section{Watermark Stealing Settings} \label{Watermark Stealing Settings}
Since mainstream watermark attack methods \cite{jovanovic2024watermark, zhang2024large, 
sadasivan2024can, gu2024on,
luo2024lostoverlapexploringwatermark, pang2024no} primarily target the red-green word list approach rather than the sampling method, we follow \citet{jovanovic2024watermark} to conduct a watermark-stealing attack, assuming the attacker has access to the distribution of unwatermarked tokens. In this attack, we query the watermarked LLM to generate a total of 200k tokens, estimate the watermark pattern, and subsequently launch spoofing attacks based on the estimated pattern. 

Specifically, we use watermarked text generated from the C4 dataset to learn the watermark, then execute a watermark spoofing attack on Dolly-CW datasets \cite{conover2023free} containing 100 samples. To ensure experimental fairness, the logits-based watermark in our hybrid symbiotic watermark employs the Unigram algorithm with identical hash keys and parameters $\gamma=0.25$, $\delta=0.4$. For the sampling-based watermark, we utilize the AAR \cite{Aaronson} algorithm. We use LLaMA2-7B-chat-hf as both the watermark and attack model, with the watermark spoofing strength set to 5.0. All other parameter settings remain consistent with those in our main experiment. 

During the watermark detection stage, we set the spoofing watermark z-score threshold to 6 and apply the original KGW watermark detection algorithm to analyze $n$ spoofing samples. If the computed z-score exceeds 6, the attack is deemed successful; otherwise, it is considered unsuccessful. Consequently, the attack success rate (ASR) is determined as follows:
\begin{equation}
    \text{ASR}=\frac{1}{n}\sum_{i=1}^n\mathbbm{I}[\text{z-score}_i>6]
\end{equation}

\end{document}